# On the Testability of Causal Models with Latent and Instrumental Variables

Judea Pearl
Cognitive Systems Laboratory
Computer Science Department
University of California, Los Angeles, CA 90024
judea@cs.ucla.edu

## Abstract

Certain causal models involving unmeasured variables induce no independence constraints among the observed variables but imply, nevertheless, inequality constraints on the observed distribution. This paper derives a general formula for such inequality constraints as induced by instrumental variables, that is, exogenous variables that directly affect some variables but not all. With the help of this formula, it is possible to test whether a model involving instrumental variables may account for the data, or, conversely, whether a given variable can be deemed instrumental.

Key words: causal modeling, instrumental variables, structural models, graphical models.

## 1 INTRODUCTION

It is well known that one cannot infer causation from statistical data unless one is willing to supplement the data with causal assumptions. Conversely, if one desires to test whether a given causal model is valid, statistical data obtained by passive (nonexperimental) observations can only refute models that constrain the joint distribution of the observables. Causal models represented by complete graphs, for example, cannot be refuted by statistical data, whereas any incomplete graph is subject to empirical falsification through the conditional independence relations induced by the missing edges. When all variables in a causal model are observable, conditional independence relationships are sufficient indeed for capturing *all* the constraints that the model imposes on the joint distribution [Verma & Pearl 1991].

This is not the case when the model invokes unobserved variables, also called *hidden* or *latent* variables. Verma and Pearl give an example where two graphs imply the same set of conditional independence relationships among the observed variables and yet they are empirically distinguishable because they imply different functional constraints on the distribution of those variables. Another such example is presented in Figure 1, where models (a) and (b) both induce no independence constraints on the observed variables, due to the spurious dependencies induced by the latent variable $U$. We shall see, however, that model (b), unlike (a), has testable implications which, if violated, can be used to falsify the model. In other words, the spurious dependencies induced by the latent variable $U$ are incapable of hiding all structural features of model (b). In contrast, model (a) is compatible with any joint distribution of $X$, $Y$, and $Z$.

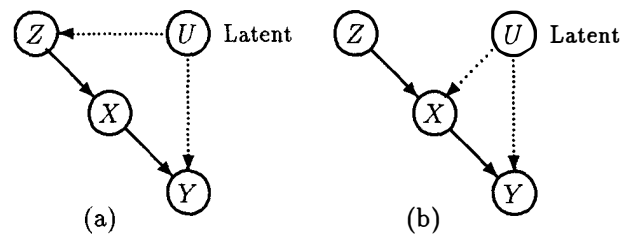

Figure 1: *Two models that induce no independence constraints on the observed variables $X, Y$, and $Z$. Model (b), unlike (a), has testable implications.*

This paper explores using the constraints induced by such structures as a mean for testing models with latent variables. The feature that makes model (b) falsifiable is the presence of two observed variables, $Z$ and $Y$, such that $Z$ is a root node and all directed paths from $Z$ to $Y$ are intercepted by observed variables. This feature constitutes a graphical definition of the notion of *instrumental variables*, which plays an important role in econometric modeling [Bowden & Turkington 1984] and in randomized trials [Imbens & Angrist 1994, Balke & Pearl 1994b]

Instrumental variables is a technique invented by the geneticist Sewal Wright [1928] to help economists identify elasticities of supply and demand [Goldberger 1972]. The key idea can be illustrated using a simple causal model given by the linear equation $y = bx + u$ in which $X$ and $Y$ are observed and $U$ represents a disturbance term, that is, unobserved factors that the modeler decides to keep out of the analysis.[1] It is well known that the coefficient $b$ in the equation above cannot be estimated consistently if $X$ and $U$ are correlated. However, if we can find a third variable $Z$ that is correlated with $X$ and is (judged to be) uncorrelated with $U$, then $b$ can be determined from the correlations between $Z$, $X$, and $Y$, yielding $b = R_{yz}/R_{xz}$. (This can be verified easily by multiplying both sides of the equation by $Z$ and taking expectations.)

Based on this simple idea, economists have developed elaborate techniques for estimating parameters in systems of linear simultaneous equations [Bowden & Turkington 1984]. More recently, the importance of this idea has grown as a consequence of the realization that some of the power of these techniques extends to nonlinear and nonparametric models, as characterized by the structural equations[2]

$$\begin{aligned} x &= g(z, u) \\ y &= h(x, u) \end{aligned} \quad (1)$$

together with the assumption that $Z$ and $U$ are independent. The two-stage process defined by these equations, which we shall name the *instrumental process*, governs almost every experimental study and it is characterized by the graph of figure 1(b). In clinical trials, for example, $Z$ represents the treatment assigned to a subject, $X$ represents the treatment actually received by a subject, and $Y$ represents an outcome of the treatment (e.g., recovery or performance). $X$ differs from $Z$ because some subjects, especially those reacting adversely to treatment, do not comply with the assignment and switch over to a different treatment group. The difficulty in analyzing such trials stems from the strong dependency between compliance and potential treatment benefit; hence, the appearance of latent variable $U$ in both equations.[3] In general, the instrumental model, as defined in Eq. (1), governs many evaluation studies in which $Z$ is a randomized instrument that encourages participation in the various programs under study, and $E_u[h(x, u)]$ represents the average merit of the program corresponding to $X = x$. Likewise, this model governs the behavior of physical systems which are subject to random influences $(U)$ and where it is required to estimate the effect of $X$ on $Y$, and only partial control over the input variable $(X)$ is possible.

The notion of instrument variables can also benefit structure-learning programs. By incorporating new variables, not directly relevant to the phenomenon under study, we can improve the reliability of structuring decisions [Pearl & Verma 1991,

Spirtes et al. 1993]. For example, if we are studying the effect of smoking $(X)$ on lung cancer $(Y)$ we might benefit from including in the analysis a variable $Z$ such as "anti-smoking legislation" which acts as an instrument for $X$ (see Figure 1(b)). In the absence of $Z$, one cannot decide whether the edge between $X$ and $Y$ should be oriented $X \to Y, X \leftarrow Y$ or eliminated altogether. The joint distribution $P(x, y, z)$, on the other hand, might supply this information, partly via conditional independencies and partly via its bare magnitude.

More ambitiously, given the model shown in Figure 1(b), one might ask whether the presence of the instrumental variable $Z$ can facilitate the identification of the causal effect of $X$ on $Y$, $E_u[h(x, u)]$, which is the nonparametric analogue of the coefficient $b$ in the linear equation $y = bx + u$. Imbens and Angrist [1994] have shown that, in general, this identification is not possible without making additional assumptions about the functions $g$ and $h$. Balke and Pearl [1994 a,b] have nevertheless shown that, it is possible to obtain sharp, informative bounds on $E_u[h(x, u)]$ without making such assumptions.

However, the question of whether a given data set can be generated by the model of Eq. (1) remains unsettled. Economists and social scientists have frequently remarked upon the difficulty of knowing or demonstrating that a variable $Z$ is instrumental, in the sense of being uncorrelated with the disturbance $U$ [Bartels 1991]. Imbens and Angrist [1994], for example, explicitly state that the model in Eq. (1) is not testable even when $Z$ is randomized. Indeed, the basic assumption embodied in the model -- that $Z$ is independent of the disturbance term of $Y$ (i.e., all factors affecting $Y$, except $X$) -- is equivalent to what economists call *exogeneity*.[4] For a long time, whether a variable $Z$ is exogenous has been thought impossible to verify experimentally, since the definition involves unobservable factors such as those represented by $U$. The notion of exogeneity, like that of causation itself, has been viewed as a subjective modeling assumption, not as an objective property that can be tested from the data.

This paper tells a different story: it shows that, despite its elusive nature, exogeneity, hence, "instrumentality," can be given an empirical test. The test is not guaranteed to detect all violations of exogeneity, but it can, in certain circumstances, screen out very bad would-be instruments.

## 2  THE INSTRUMENTAL INEQUALITY

**Definition 1** (instrument) *A variable $Z$ is said to be an instrument relative to an ordered pair of variables*

---

[1] We uses upper case symbols for variable names and lower case symbols for specific realizations of the variables.

[2] See appendix for the relationships among graphs, structural equations and counterfactuals.

[3] These equations are normally written using several "error" terms, for example, $x = g(z, \epsilon_x)$, $y = h(x, \epsilon_y)$, with $\epsilon_x$ and $\epsilon_y$ being jointly independent of $Z$. This formulation is equivalent to Eq. (1), since we can define $U$ to consist of the joint space of $\epsilon_x$ and $\epsilon_y$.

[4] This assumption is termed *superexogeneity* in [Engle, et al. 1984].

$(X, Y)$ if $X$ and $Y$ are generated by the following process:

$$x = g(z, u)$$
$$y = h(x, u) \quad (2)$$

*where $g$ and $h$ are arbitrary deterministic functions, and $U$ is an arbitrary, unobserved random variable, independent of $Z$.*

The required independence between $Z$ and $U$ rules out the possibility that $Z$ is influenced by some latent cause that also influences other variables in the system. The exclusion of $z$ from $h(\cdot)$ rules out $Z$ having any effect on $Y$ that is not mediated by $X$, thus capturing the notion of locality, whereby an instrument is presumed to "affect $X$ only." Note that no restrictions are posed on the domain of $U$; it may be finite or unbounded, discrete or continuous, ordered or unstructured.

Our problem is to determine from the observed joint probability distribution $P(x, y, z)$ whether $Z$ can be exogenous relative to $(X, Y)$, that is, whether there exist two functions $g$ and $h$ and a probability distribution on $U$ and $Z$ (with $Z$ and $U$ independent) such that the distribution generated by the two equations corresponds precisely to the observed distribution $P(x, y, z)$.

**Theorem 1** *A necessary condition for discrete variables $X, Y$ and $Z$ to be generated by an instrumental process as defined in Eq. (2) is that the conditional distribution $P(x, y|z)$ satisfies*

$$\max_x \sum_y [\max_z P(x, y|z)] \leq 1 \quad (3)$$

**Proof** If the probability distribution $P(x, y, z)$ is generated by the process defined in Eq. (2), then it can be expressed in the form

$$P(x, y, z) = \sum_u P(y|x, u) P(x|z, u) P(u) P(z)$$

This can be seen by decomposing $P(x, y, z, u, v)$ into product form along the order $(y, x, v, u, z)$ and using the independence relations imposed by the model of Eq. (2), as displayed in the graph of Figure 1(b). Therefore,

$$\begin{aligned} P(x, y|z) &= \sum_u P_1(y|x, u) P_2(x|z, u) P(u) \\ &= E_u [P(y|x, u) P(x|z, u)] \end{aligned} \quad (4)$$

If Eq. (4) holds for every triplet $(x, y, z)$, it certainly holds for a select set of triplets $(x, y, z(x, y))$, where $z(x, y)$ is chosen so as to maximize $P(x, y|z)$. Thus, summing Eq. (4) over $y$, gives

$$\sum_y P(x, y|z(x, y)) = E_u \sum_y P(y|x, u) P(x|z(x, y), u) \quad (5)$$

For any fixed $x$ and $u$, the term $P(x|z(x, y), u)$ can be considered a function of $y$, which is bounded from above by unity. The summation on the r.h.s. of Eq. (5) represents a convex sum of such $P$ terms and, hence, it must also be bounded by unity, which, after substituting out $z(x, y)$, gives

$$\sum_y \max_z P(x, y|z) \leq 1 \quad (6)$$

Moreover, since this inequality must hold for every $x$, we can write

$$\max_x \sum_y [\max_z P(x, y|z)] \leq 1 \quad (7)$$

which proves the theorem. □

We call the inequality above an *instrumental inequality* because it constitutes a necessary condition for any variable $Z$ to qualify as an instrument relative to $(X, Y)$. Recently, Huy Cao (1995) has shown that the inequality is not sufficient, except when $X$ is binary.

## 3   INTUITIONS, APPLICATIONS, AND EXTENSIONS

If all observed variables are binary, Eq. (7) reduces to the four inequalities

$$P(Y = 0, X = 0|Z = 0) + P(Y = 1, X = 0|Z = 1) \leq 1$$
$$P(Y = 0, X = 1|Z = 0) + P(Y = 1, X = 1|Z = 1) \leq 1$$
$$P(Y = 1, X = 0|Z = 0) + P(Y = 0, X = 0|Z = 1) \leq 1$$
$$P(Y = 1, X = 1|Z = 0) + P(Y = 0, X = 1|Z = 1) \leq 1 \quad (8)$$

which were derived in an analysis of noncompliance in experimental studies [Pearl 1993].

We see that the instrumental inequality is violated when the controlling instrument $Z$ manages to produce significant changes in the response variable $Y$ while the direct cause, $X$, remains constant. Although such changes could in principle be explained by spurious correlation through $U$, since $X$ does not screen off $Z$ from $Y$, the instrumental inequality sets a limit on the magnitude of the changes. The similarity to Bell's inequality in quantum physics [Cushing & McMullin 1989, Suppes 1988] is not accidental; both inequalities delineate a class of observed correlations that cannot be explained by hypothesizing latent common causes. The instrumental inequality can, in fact, be viewed as a variant of Bell's inequality for cases where direct causal connection is permitted to operate between the correlated observables $X$ and $Y$.

Of special interest to experimenters is the prospect of applying the instrumental inequality to the detection of undesirable side-effects in experimental studies. In clinical trials, for example, dependencies between the treatment assignment ($Z$) and factors ($U$) affecting the response process can be attributed to one of two possibilities: either there is a direct causal effect of the assignment ($Z$) on the response ($Y$), unmediated by the treatment ($X$), or there is a common causal factor correlating the two ($Z$ and $U$). If the assignment is carefully randomized, then the latter possibility is ruled out and any violation of the instrumental inequality (even under conditions of imperfect compliance) can safely be attributed to some direct influence of the assignment process on subjects' response (e.g., psychological aversion to being treated). Alternatively, if one can rule out any direct effects of $Z$ on $Y$, say through effective use of a placebo, then any observed violation of the instrumental inequality can safely be attributed to spurious dependence between $Z$ and $U$, namely, to selection bias.

The instrumental inequality can be tightened appreciably if we are willing to make additional assumptions about subjects' behavior – for example, that no individual can be discouraged by the encouragement instrument, or, mathematically, that for all $u$ we have

$$g(z_1, u) \geq g(z_2, u)$$

whenever $z_1 \geq z_2$. Such an assumption amounts to having no contrarians in the population, namely, no individual who would consistently act contrary to his or her assignment. Under this assumption, which Imbens and Angrist [1994] call monotonicity, the inequalities in Eq. (3) can be tightened [Balke & Pearl 1994a] to give

$$\begin{aligned} P(y, X = 1 | Z = 1) &\geq P(y, X = 1 | Z = 0) \\ P(y, X = 0 | Z = 0) &\geq P(y, X = 0 | Z = 1) \end{aligned} \quad (9)$$

for all $y \in \{0, 1\}$. Violation of these inequalities now means either selection bias or a direct effect of $Z$ on $Y$ or the presence of contrarian subjects.

## 4   THE ENIGMATIC CONTINUUM

Extending the instrumental inequality to the case where $Z$ and $Y$ are continuous presents no special difficulty. If $f(y|x, z)$ is the conditional density function of $Y$ given $X$ and $Z$, then tracing the proof of Theorem 1 gives a condition similar to Eq. (3):

$$\int_y \max_z [f(y|x, z) P(x|z)] dy \leq 1 \quad \forall x \quad (10)$$

However, the transition to continuous $X$ involves a drastic change of behavior, and it seems that Eq. (2) induces no constraints whatsoever on the observed density.

It is clear that any tri-variate normal distribution $f(x, y, z)$ can be generated by a process in which $Z$ is instrument for $(X, Y)$, as defined by Eq. (2). This can be seen from the fact that for any set of correlation parameters $R_{xy}, R_{yz}$, and $R_{zx}$ ($R_{zx} > 0$), we can always find a (unique) solution for the coefficients $\hat{a}, \hat{b}$, and $\hat{c}$ in the equations

$$\begin{aligned} x &= \hat{a} z + \hat{c} u \\ y &= \hat{b} x + u \end{aligned} \quad (11)$$

(the linear version of Eq. (2)) so as to satisfy the given correlation parameters under the assumption that $Z$ and $U$ are uncorrelated. In particular, this solution yields the celebrated *instrumental variable estimator*

$$\hat{b} = R_{zy}/R_{zx} \quad (12)$$

which may have no relation whatsoever with the process which actually governs the generation of $Y$ in the data. Thus, the instrumental inequality cannot weed out a bad instrument $Z$ if all measured variables are normally distributed.

The essential difference between the discrete and the continuous cases can be seen from the last step in the proof of Theorem 1. If in Eq. (5) we substitute the densities $f(y|x, u)$ and $f(x|z(x, y), u)$ instead of probabilities, we obtain

$$\int_y \max_z f(y|x, z) f(x|z) dy = E_u \int_y f(y|x, u) f(x|z(x, y), u) dy \quad (13)$$

However, in contrast to Eq. (5), we can not bound $f(x|z(x, y), u)$ below unity.

**Conjecture 1** *If $x$ is continuous, then every joint density $f(y, x|z)$ can be generated by the instrumental process defined in Eq. (2).*

Although we are currently only close to achieving a general proof of Conjecture 1, some interesting special cases are worth reporting here.

**Definition 2** (generator) *Given a conditional density $f(x|z)$, a function $x = g(z, u)$ is said to be a **generator** of $f(x|z)$ iff there exists some probability measure on the domain of $U$ such that $g$ is distributed as $f(x|z)$, namely, $P[g(z, u) \leq x] = F(x|z)$ when $F(x|z)$ is the cumulative conditional distribution associated with $f(x|z)$.*

**Definition 3** (one-to-one generator) *A generator $g(z, u)$ of $f(x|z)$) is said to be one-to-one iff, for every $x$ and $u$, the equation $x = g(z, u)$ has a unique solution for $z$. In other words, $g(z_1, u) = g(z_2, u)$ implies $z_1 = z_2$.*

**Lemma 1** *Any density $f(y, x|z)$ whose marginal $f(x|z)$ has a one-to-one generator can be generated by an instrumental process (Eq. (2)).*

**Proof:** If $g(z, u)$ is a one-to-one generator of $f(x|z)$, we write

$$f(y, x|z) = f(y|x, z) f(x|z)$$

then use $x = g(z, u)$ to generate $f(x|z)$, and some other function $y = h'(x, z, v)$ to generate $f(y|x, z)$, where $u$ and $v$ are independent random variables. Since $g$ is one-to-one, we can compute a unique $z = g^{-1}(x, u)$ for each pair $(x, u)$. Hence, we can substitute out $z$ from $h'(\cdot)$ and obtain

$$y = h'(x, g^{-1}(x, u), v) = h(x, u, v)$$

which conforms to the process defined in Eq. (2) if we consider $u$ as representing the pair $(u, v)$.

**Example 1** Let

$$f(y,x|z) = \begin{cases} 2x/z & 0 \leq y \leq z,\ 0 \leq x \leq 1 \\ 0 & \text{otherwise} \end{cases} \quad (14)$$

The marginal of this density is $f(x|z) = 2x$ ($0 \leq x \leq 1$), which is independent of $z$. Thus, $Z$ has no effect on $X$, yet, for any fixed $x$, $Z$ has an effect on the density of $Y$, because

$$f(y|x,z) = 1/z \quad 0 \leq y \leq z$$

By constructing a one-to-one generator for $f(x|z)$, we will show that $f(y,x|z)$ can still be generated by an instrumental process, in which $Z$ has no direct effect on $Y$.

The method of constructing such a generator was shown to me by Steffen Lauritzen (personal communication, January 1995). We first define a new variable $X'$ as

$$x' = (z+u)\bmod(1) \stackrel{\Delta}{=} z \oplus u \quad (15)$$

and let $U$ be distributed uniformly over $[0,1]$. Clearly, the distribution of $x'$ is uniform over $[0,1]$ for all values of $z$; moreover, $\oplus$ has a unique inverse for $z \in [0,1]$, which we write $z = x' \ominus u$. We now express $x$ as a function of $x'$ so as to endow $x$ with the desired conditional density $f(x|z) = 2x$ (i.e., $F(x|z) = x^2$). The proper transformation is $x = F^{-1}(x'|z) = \sqrt{x'} = \sqrt{z \oplus u}$. This defines a generator for $f(x|z)$,

$$x = g(z,u) = \sqrt{z \oplus u} \quad (16)$$

which is one-to-one, because we can invert this equation to obtain

$$z = g^{-1}(x,u) = x^2 \ominus u$$

We are now in a position to construct the conditional density $f(y|x,z) = 1/z, 0 \leq y \leq z$, by letting $y$ be a function $h$ of $x$ and $u$. First, we construct the desired density by the standard method, letting $y$ be a function of $x, z$, and $v$,

$$y = F_Y^{-1}(v|x,z) = vz \quad (17)$$

where $v$ is uniformly distributed over $[0,1]$, independent of $u$. Second, we substitute out $z$ and obtain

$$y = h(z,u) = v(x^2 \ominus u)$$

which, together with Eq. (16), generates the joint density specified in Eq. (14).

**Corollary 1** *Every density $f(y,x,z)$ satisfying $f(x|z) = f(x)$ (i.e., $Z$ and $X$ are independent) can be generated by the instrumental process of Eq. (2).*

This can be seen by generalizing the construction of Lemma 1 to arbitrary density $f(x|z)$. To this end we again define $x' = z \oplus u$, and let $u$ be distributed uniformly over $[0,1]$. In order to insure the proper density on $x$, we invoke the transformation

$$x = F^{-1}(x'|z)$$

where $F^{-1}(x'|z)$ stands for the inverse cumulative distribution associated with $f$ (i.e., $x' = F(x|z)$). Thus, our overall generator is

$$x = g(z,u) = F^{-1}(u \oplus z|z) \quad (18)$$

To complete the construction, it is sufficient to show that this equation has a unique solution for $z$, which is certainly the case whenever $F(x|z) = F(x)$.

Although this construction does not apply to a general $f(x|z)$, it enabled Huy Cao [1995] to prove Conjecture 1 for the case of $Z$ is countably discrete $Z$.

**Observation** If $x$ is discrete, then $f(x|z)$ may not have a one-to-one generator. In particular, the existence of a one-to-one generator is categorically ruled out if the corresponding probability mass function $P(x|z)$ satisfies

$$p(x|z_1) + p(x|z_2) > 1$$

for some $x, z_1$, and $z_2$. In other words, every generator $g(z,u)$ of such a $P(x|z)$ has some $u$ for which both $z_1$ and $z_2$ are mapped into the same $x$ and, moreover, the set $U_0$ of such $u$'s must have a nonzero probability.

**Remark** To appreciate the importance of one-to-oneness, we can think of $z$ as an action that is applied to some population, $u$ as denoting a given unit in the population, and $x = g(z,u)$ as the response of unit $u$ to action $z$. If $x$ is discrete, then whenever we observe

$$P(x|z_1) + P(x|z_2) > 1$$

we may conclude that a non-negligible fraction of the population must be nonresponsive to the action, that is, $g(z_1,u) = g(z_2,u)$ for every $u$ in that subpopulation, where $z_1$ and $z_2$ are two different actions. In contrast, no observation on continuous densities would imply the existence of such nonresponsive subpopulations because every density $f(x|z)$ can be realized by a population in which the response $x$ of every individual is always sensitive to variations in actions, namely,

$$\forall u \quad g(z_1,u) \neq g(z_2,u) \text{ if } z_1 \neq z_2 \quad (19)$$

Even the extreme case of $f(x|z) = f(x)$ can be realized in a fully responsive population satisfying Eq. (19), as is demonstrated by the one-to-one generator of Example 1.

**Pending questions** Additional questions come to mind when we restrict the functional form of the generator $g$: Does every $f(x|z)$ have a one-to-one generator if we limit our consideration to

1. smooth generators, that is, $g(z,u)$ differentiable in $z$ for every $u$;
2. monotonic generators, that is, $g(z_1,u) \geq g(z_2,u)$ whenever $z_1 > z_2$;
3. smooth and monotonic generators.

The answer, most probably, is no. Even the independent case, $f(x|z) = f(x)$, does not seem to have any smooth or monotonic one-to-one generator. Thus, the questions we need to answer are as follows:

1. What characterizes those densities $f(x|z)$ that do possess one-to-one monotonic generators.

2. What characterizes the class of densities that do not have a one-to-one monotonic generator but still can be generated using instrumental monotonic generators.

## Acknowledgment

The research was partially supported by Air Force grant #AFOSR/F496209410173, NSF grant #IRI-9420306, and Rockwell/Northrop Micro grant #94-100. Ed Leamer made helpful suggestions on the first draft of this paper. Steffen Lauritzen and Huy Cao have illuminated this topic with new results.

W.L. (Eds.), *Causation, Chance, and Credence*, 135–151, Kluwer Academic Publishers, Dordrecht, The Netherlands, 1988.

[Verma & Pearl 1991] Verma, T.S., and Pearl, J., "Equivalence and synthesis of causal models," in *Uncertainty in Artificial Intelligence*, Vol. 6, 220–227, Elsevier Science Publishers, Cambridge, MA, 1991.

[Wright 1928] Wright, P.G., *The Tariff on Animal and Vegetable Oils*, Macmillan, New York, 1928.

## APPENDIX: GRAPHS, STRUCTURAL EQUATIONS AND COUNTERFACTUALS

This paper uses two representations of causal models: graphs and structural equations. By now, both representations have been considered controversial for almost a century. On the one hand, economists and social scientists have embraced these modeling tools, but they continue to debate the empirical content of the symbols they estimate and manipulate; as a result, the use of structural models in policy-making contexts is often viewed with suspicion. Statisticians, on the other hand, reject both representations as problematic (if not meaningless) and instead resort to counterfactual notation whenever they are pressed to communicate causal information.[5] This appendix presents an explication that unifies these three representation schemes in order to uncover commonalities, mediate differences, and make the causal-inference literature more generally accessible.

**Structural models**
The natural place to start is with a system $T$ of structural equations like those in Eq. (1), which for the purposes of exposition we now write as

$$\begin{aligned} z &= f(u_Z) \\ x &= g(z, u_X) \\ y &= h(x, u_Y) \end{aligned} \qquad (20)$$

These equations describe the physical processes that generate the observed data: instances $(x, y, z)$ of variables $X$, $Y$, and $Z$. The value $z$ of $Z$ is determined by an unobserved factor $U_Z$ which we choose to keep outside the analysis. The value $x$ of $X$ is determined by two factors, the value of $Z$ and an external factor $U_X$, and so on. If the functions $f$, $g$, and $h$ are known, the model is *parametric*; otherwise, it is *nonparametric*. The *contextual* variables $U = (U_X, U_Y, U_Z)$ summarize the environment external to the system under analysis. Their values are determined outside the system, hence they are often called *exogenous*.[6] They may stand for factors such as "weather conditions" or "life style" for which we have verbal descriptions, or they simply may serve as generic symbols for *all* the factors that were omitted from the analysis. Unlike regression models, structural equations make no a priori assumptions regarding independencies among the $U$ variables.

The two defining attributes of structural equations which set them apart from ordinary algebraic equations are autonomy [Haavelmo 1943] and asymmetry [Simon 1953]. These attributes do not show up explicitly, as symbols in the equations, but implicitly, by attaching meaning to any *subset* of equations from $T$, thus restricting the type of algebraic transformations that are semantic-preserving.

Autonomy reflects the understanding that the three equations above represent three independent processes, and hence the three equations in (20) convey more information than the single vector mapping

$$v \triangleq (x, y, z) = F(u_X, u_Y, u_Z) \triangleq F(u) \qquad (21)$$

which would be obtained from Eq. (20) by substitution. Indeed, for a given value of $u$, Eq. (21) provides merely a point value for $(x, y, z)$, while Eq. (20) also provides information about how that point value will change under a class of interventions which alter a selected subset of equations.[7] Eq. (20) tells us that, no matter what changes are made in the values of $X$, $Z$, and $U$, or in the processes $(f, g)$ generating $z$ and $x$, the value of $Y$ will remain $h(x, u_Y)$ whenever $X$ and $U_Y$ take on the values $x$ and $u_Y$, respectively. The empirical content of this information can be operationalized using a hypothetical experiment in which the values of $X$ and $U_Y$ are held constant by some external control. Under such conditions, Eq. (20) predicts the relation $y = h(x, u_Y)$ to hold permanently, irrespective of any controls applied to other variables in the system.

Asymmetry reflects the directionality of the relationship "is determined by." The equation $y = h(x, u_Y)$ prescribes how changes in $x$ (be they a product of new interventions or of changes in $u$) would translate into changes in $y$, but not the other way around. Thus, the equality sign in structural equations models is not an ordinary algebraic equality but functions more like the assignment symbol in programming languages. The identity of the dependent variable (positioned on the lhs) in each equation is useful for indexing the equations to be altered by a given intervention, say, holding $X$ fixed.

---

[5] Space limitations do not permit us to offer an elaborate account of these formulations or to provide references to history of these controversies. For a more detailed account, see [Pearl 1995] and the references cited therein.

[6] This notion of exogeneity (synonymous with predeterminedness) is much weaker than that used in the text; the boundary between contextual and endogenous variables is often a matter of modeling choice and does not rest on any assumption of independence relative to error terms.

[7] Unlike most of the literature (e.g., [Simon 1953]) we do not insist that interventions be represented as changes in $U$; modellers need not anticipate in advance all interventions capable of altering a given process.

These process-based considerations are important in the modeling phase, when the structural equations are put together. Once completed, causal analysis proceeds on the basis of syntactic structure alone. Given a set $T$ of structural equations, one can define complex notions such as causation, intervention, atomic intervention, causal effect, causal relevance, average causal effect, plans, conditional plans, identifiability, counterfactuals, exogeneity, and so on. For example, the atomic intervention $set(X = x)$ is modeled by replacing the equation corresponding to $X$ with the equation $X = x$ and then solving the resulting set of equations for the variables of interest [Strotz & Wold 1971]. Accordingly, we can say that "$X$ is a cause of $Y$ in context $u$" if there are two values of $X$, $x$ and $x'$, such that the solution for $Y$ under $U = u$ and $set(X = x)$ is different from the solution under $U = u$ and $set(X = x')$.

Probabilistic causality emerges when we define a probability distribution $P(u)$ for the $U$ variables, which, under the assumption that the equations have a unique solution, induces a unique distribution on the endogenous variables for each combination of atomic interventions. The causal effect of $X$ on $Y$, denoted $P(y|\hat{x})$, is then defined as the distribution of $Y$ induced by $P(u)$ under the intervention $set(X = x)$ [Pearl 1995]. Recently, wealth of new results have been obtained on nonparametric identification of $P(y|\hat{x})$, thus providing conditions under which $P(y|\hat{x})$ depend not on the functions $f(\cdot), g(\cdot), h(\cdot), \ldots$, but only on the observed distributions (see [Galles & Pearl 1995, Pearl & Robins 1995] in this volume). These conditions require that $P(u)$ exhibit a rich set of independencies and that the equations be sparse and recursive. The precise specification of these requirements is best expressed in terms of graphs.

### Graphs

Graphs offer an abstraction of structural equations. They carry the following two pieces of information:

1. The identity of the observable variables on the rhs of each equation (often called *independent variables*). These are represented as nodes in the graph from which arrows emanate into the dependent variable in the equation. Thus, each equation translates into a *parents-child family* in a directed graph. The directed graph may be either cyclic, or in the case where the equations are recursive, acyclic. The parents of variable $X$ will be denoted by $\Pi_X$, and any realization of those parent variables by $\pi_X$.

2. The identity of jointly independent groups of $U$ terms in $P(u)$. This information is represented graphically as double-arrow dashed arcs between pairs of nodes. The absence of a dashed arc between node $X$ and a set of nodes $Z_1, \ldots, Z_k$ implies that the corresponding $U$ variables, $U_X, U_{Z_1}, \ldots, U_{Z_k}$, are jointly independent.[8]

---

[8]In principle, there may be several sets of arcs repre-

For example, the graph in Figure 1(b) represents the structural equations in (20), since each parents-child family in the graph corresponds to one equation in (20), with parent sets:

$$\Pi_Z = \{\emptyset\}, \ \Pi_X = \{Z\}, \ \Pi_Y = \{X\}$$

In addition, the absence of a dashed arc between $Z$ and $X$ and between $Z$ and $Y$ represents the independence $U_Z \parallel \{U_X, U_Y\}$ or, equivalently, $Z \parallel \{U_Z, U_Y\}$.

### Counterfactuals

The counterfactual notation, usually associated with Rubin's model [Rubin 1974] (some economists refer to it as Roy's model), represents another abstraction of structural equations. The starting point is not the system of equations but the set of solutions of those equations under different contexts (or *units*) $U = u$.[9] The primitive object of analysis is the unit-based response variable, denoted $Y(x, u)$ or $Y_x(u)$, which stands for the solution for $Y$ under $U = u$ and under the hypothetical intervention $set(X = x)$ ($X$ may stand for a subset of variables).[10]

To statisticians, the attractive feature of the counterfactual notation is that it permits prior causal knowledge to be expressed as assumptions about random variables (albeit counterfactual), thus allowing the analysis to remain within the boundaries of standard probability calculus. This conforms fully with the Bayesian requirement that all prior knowledge be expressed as constraints on distributions; meta-probabilistic notions such as "exogeneity," "processes," "autonomy," and "intervention" are avoided, at least superficially.

If $U$ is treated as a random variable, then the value of the counterfactual $Y(x, u)$ becomes a random variable as well, denoted as $Y(x)$ or $Y_x$. Causal analysis can then proceed by imagining the observed distribution $P(x, y, z)$ as the marginal distribution of an augmented probability function $P^*$ defined over both the observed variables and the counterfactual variables of

---

senting the same independencies in the manner described above. However, if the dependencies among the $U$ variables are themselves a product of a recursive causal process, the arc representation is unique.

[9]The term *unit* instead of *context* is used in the counterfactual literature [Rubin 1974], where it normally stands for the identity of a specific individual in a population, namely, the set of attributes that characterize that individual. This is precisely the role played by the vector $u$ in structural equations. In general, $u$ may include the time of day, the experimental conditions under study, and so on.

[10]Practitioners of the counterfactual notation do not explicitly mention the notions of "solution" or "intervention" in the definition of $Y(x, u)$. Instead, the phrase "the value that $Y$ would take in unit $u$, had $X$ been $x$," viewed as basic, is posited as the definition of $Y(x, u)$. However, since structural models offer a formal semantics (based on solving subsets of equations) for counterfactual phrases of this kind, the definition above is deemed more basic, and it helps illuminate the connection between structural models and counterfactual variables.

interest, say $Y(x)$. Queries about causal effects, previously written $P(y|\hat{x})$, are rephrased as queries about the marginal distribution of the counterfactual variable of interest, written $P(Y(x) = y)$. The new entities $Y(x)$ are treated as ordinary random variables that are connected to the observed variables via the logical constraint [Robins 1987]

$$X = x \implies Y(x) = Y \qquad (22)$$

and a set of conditional independence assumptions which the investigator must supply to endow the augmented probability, $P^*$, with causal knowledge. These assumptions should encode (a summary of) the investigator's understanding of the data-generation process, previously encoded in equations or in graphs.

For example, to convey the understanding that the process generating $X$ in Eq. (20) is not in itself affected by the variable $Z$, the analyst should communicate the independence constraint $X(z) \perp\!\!\!\perp Z$. Likewise, to communicate the understanding that in a randomized clinical trial the way subjects react $(Y)$ to treatments $(X)$ is statistically independent of the treatment assignment $(Z)$, the analyst would write $Y(x) \perp\!\!\!\perp Z$.

A collection of constraints of this type might sometimes be sufficient to permit a unique solution to the query of interest, e.g., $P(Y(x) = y)$; in other cases, only bounds on the solution can be obtained. The models in Figure 1(a) and 1(b) are examples of the former and latter cases, respectively. It should be remarked though, that since users of the counterfactual notation do not view counterfactual variables as by-products of a deeper model of the data-generating mechanism, the process of issuing judgments about counterfactual dependencies has not been systematized. Analysts are not always sure whether *all* relevant judgments have been articulated, whether the judgments articulated are redundant, or whether those judgments are consistent with the data. Such judgments can be systematized in the graphical representation of structural equations, as is shown next.

**Translation: From Graphs to Counterfactuals**

The assumptions embodied in a causal graph can be translated into the counterfactual notation using two simple rules; the first interprets the missing arrows in the graph, the second, the missing dashed arcs. Missing arrows represent variables that were deemed excludable from (the process or the hypothetical experiment described by) an equation. Missing arcs encode independencies among the $U$ terms in two or more equations.

1. Exclusion restrictions: For every variable $Y$ having parents $\Pi_Y$, and for every set of variables $S$ disjoint of $\Pi_Y$, we have

$$Y(\pi_Y) = Y(\pi_Y, s) \qquad (23)$$

2. Independence restrictions: For every pair of variables $X$ and $Y$ not connected by a dashed arc,

we have

$$Y(\pi_Y) \perp\!\!\!\perp X(\pi_X) \qquad (24)$$

Likewise, if $X_1, \ldots, X_k$ is any set of nodes not connected to $Y$ via dashed arcs, we have

$$Y(\pi_Y) \perp\!\!\!\perp \{X_1(\pi_{X_1}), \ldots, X_k(\pi_{X_k})\} \qquad (25)$$

For example, the graph in Figure 1(a), displaying the parent sets

$$\Pi_Z = \{\emptyset\}, \Pi_X = \{Z\}, \Pi_Y = \{X\}$$

encodes the following assumptions:

1. Exclusion restrictions:

$$Z(x) = Z(y) = Z(x, y) = Z(\emptyset) \stackrel{\Delta}{=} Z$$
$$X(y, z) = X(z), Y(x) = Y(x, z)$$

2. Independence restrictions:

$$X(z) \perp\!\!\!\perp \{Z, Y(x)\}$$

We leave it to the reader to show that these assumptions are sufficient to allow computation of the causal effect $P(Y(z) = y)$ using standard probability calculus together with axiom (22). Note that, unlike the independence judgments normally required by counterfactual analysts, the assumptions obtained from the graph involve only parents-child relationships (e.g., $Y(x), X(z)$); remotely related counterfactuals such as $Y(z)$, which are cognitively less meaningful, may be derived from parents-child relationships by substitutions but are not the object of direct subjective judgments.

It is also interesting to note that the analysis of instrumental inequalities presented in this paper is valid under more general conditions than those shown in the graph of Figure 1(b). If an arrow from $Y$ to $X$ is added to the graph, a cyclic graph containing the feedback loop $X \to Y \to X$ is obtained. (In the context of clinical trials, such a loop may represent, for example, patients deciding on dosage $X$ by continuously monitoring their response $Y$.) Nonetheless, the structural equation model will not change, because, under the assumption that the cycle is stable, the equation

$$x = g(z, y, u)$$

can be replaced with

$$x = g'(z, u')$$

such that $u'$ is still independent of $z$. The nonparametric nature of the structural equations in (20) permit us to make such transformations without affecting the results of the analysis. Likewise, nonparametric bounds obtained from the analysis of the acyclic graph 1b [Balke & Pearl, 1994a,b] are still valid for the cyclic case.